# Motion Control Simulations for a Magnetically-Coupled Bacterium and Robotic Arm

Ahmet Fatih Tabak, *Member*, IEEE
Mechatronics Engineering Department
Bahcesehir University
Istanbul, Turkey
ORCID: 0000-0003-3311-6942

*Abstract*—The demonstrations of micro-robotic systems in minimally invasive medicine include an individual or a swarm of microswimmer of various origin, artificial or biohybrid, often with an external computer-controlled electromagnetic field. There are several *in vivo* and *in vitro* control performances with artificial microswimmers but control of a bio-hybrid microswimmer using an open kinematic chain remains untouched. In this work, non-contact maneuvering control of a single magnetotactic bacterium cell is simulated. The results show that the proposed system is capable of adjusting the heading of the microswimmer moving at proximity to a 2D boundary under the guidance of the set-point tracking scheme. The performance of the coupled model and the sensitivity to control parameters are demonstrated with the help of a time-dependent error to the yaw-angle reference under the influence of PID with adaptive integral gain.

## I. INTRODUCTION

The use of microrobots of different origins in possible therapeutic applications is widely demonstrated in the literature [1] – [5]. There are acoustic, optic, chemical, biohybrid, and magnetic microrobots, single or in a swarm, used in in vivo and in vitro demonstrations [6] – [14]. The magnetic microrobots, especially of bioinspired type, further can be categorized as artificial, biohybrid, and magnetotactic [15] – [17]. Motion control of such systems is widely dependent on the electromagnetic field and respective field gradients generated by computer-controlled electric currents flowing through specially-arranged electromagnetic coils [18], [19]. The coils are either of an MRI system, or a smaller and stationary system, or mounted at the end-effector of a robotic arm [20] – [22]. Furthermore, open- and closed-loop control efforts have recently proliferated in the literature [23] – [36]. However, the control of magnetotactic bacteria via permanent magnets at the end-effector of an open-kinematic chain is not widely studied to date.

The electromagnetic coils have the two immediate disadvantages: (i) apparent heat generation, hence the cooling problem accompanied with high current demand for strong fields; (ii) the field generated by an electromagnetic coil setup has a limited volume of interest with homogenous field or gradient to actuate the magnetic microrobot [37]-[39]. The microrobot is mostly placed in the middle of the coil system, although some different successful specialized designs are available for relatively small workspaces [40]. However, if one wants to track the microrobot along with a relatively longer gait in living tissue, either the coil system should be larger, such as an MRI system, or the electromagnetic field should follow the microrobot. One possible solution is to follow the microrobot, which is a natural microswimmer propelling itself without external power input, via a permanent magnet and such a task could be achieved by a robotic arm.

In this study, a detailed mathematical model comprising of coupled equations of an open kinematic chain and a microswimmer is built along with a semi-adaptive PID controller achieving a bilateral control. Results verify that such a system can steer a magnetotactic bacterium cell, which will be referred to as microswimmer hereon. The rest of the paper deals with the complete mathematical model of the proposed system in great detail and provides comparative results of the simulations over different control parameters.

## II. MATHEMATICAL MODEL OF THE SYSTEM

The proposed system is the combination of a single individual of a microswimmer species (*M. Gryphiswaldense*) [41] with but one helical bundle for the tail, one Prismatic-Prismatic-Revolute (PPR) arm with the third link being embedded in the second one to simplify the dynamics, and an N52-grade Neodymium magnet [42] of rectangular-prism shape (see Fig.1) as such even without the control the magnetic interaction would result in some forced-orientation on the swimming direction of the microswimmer. However, as explained in this section, the control law complements the two-way coupling such that the two systems act as one.

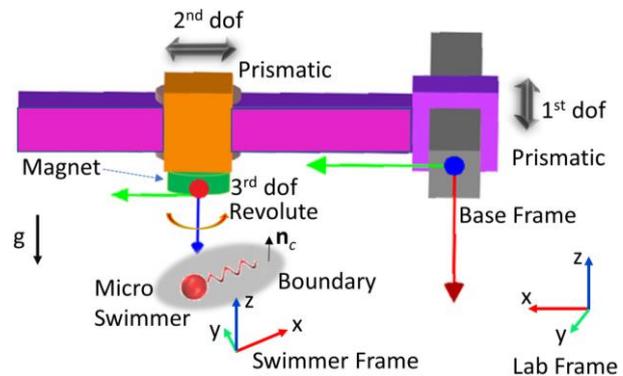

Fig. 1. Two robots placed in the lab frame: The PPR arm and the micro swimmer. The PPR robot visual is obtained via RoboAnalyzer [43].

The author would like to thank Istanbul Commerce University for the MATLAB (MathWorks®) license used for the analysis in this study.

The equation of motion cast for the microswimmer takes several physical stimuli as follows:

$$\begin{bmatrix} \mathbf{F}_p(t)+\mathbf{F}_d(t)+\mathbf{F}_m(t)+\mathbf{F}_g(t)+\mathbf{F}_c(t) \\ \mathbf{T}_p(t)+\mathbf{T}_d(t)+\mathbf{T}_m(t)+\mathbf{T}_g(t)+\mathbf{T}_c(t) \end{bmatrix} = \mathbf{0}. \quad (1)$$

Here, the vectors $\mathbf{F}_p$ and $\mathbf{T}_p$ represent the 6-dof propulsive effect associated with the rotating flagellar bundle of the microswimmer. The model is based on resistive force theory [44] (RFT) and constructed as:

$$\begin{bmatrix} \mathbf{F}_p \\ \mathbf{T}_p \end{bmatrix} = \left( \int_0^L \begin{bmatrix} \mathbf{RCR}' & -\mathbf{RCR}'\mathbf{S} \\ \mathbf{SRCR}' & -\mathbf{SRCR}'\mathbf{S} \end{bmatrix} d\ell \right) \begin{bmatrix} \mathbf{0} \\ \mathbf{\Omega}_{tail} \end{bmatrix}, \quad (2)$$

where $\mathbf{R}$ denotes the local Frenet-Serret rotation, $\mathbf{S}$ is the local cross-product matrix, and $\mathbf{C}$ gives the local fluid resistance matrix taking the hydrodynamic interactions and transient effects into account [45] to make the analysis as realistic as possible. Furthermore, all these matrices are 3×3 and expected to be subject to temporal and spatial changes along the tail due to the helical wave propagation in a time-irreversible manner close to a solid boundary [46]. Thus the integral is taken along the entire tail. Since the integral reveals a 6×6 full matrix [45] the rotation of the helical tail, i.e., $\mathbf{\Omega}_{tail} = [\Omega_x\ 0\ 0]'$, results in force and torque components in all main axes. The main opposing effect is the fluidic drag which is modeled as:

$$\begin{bmatrix} \mathbf{F}_d \\ \mathbf{T}_d \end{bmatrix} = -\left\{ \int_0^L \begin{bmatrix} \mathbf{RCR}' & -\mathbf{RCR}'\mathbf{S} \\ \mathbf{SRCR}' & -\mathbf{SRCR}'\mathbf{S} \end{bmatrix} d\ell \right. \\ \left. + \begin{bmatrix} \mathbf{D} & -\mathbf{DS}_{body} \\ \mathbf{S}_{body}\mathbf{D} & \mathbf{E} \end{bmatrix} \right\} \begin{bmatrix} \mathbf{U} \\ \mathbf{\Omega} \end{bmatrix}_{sw}. \quad (3)$$

In (3), $\mathbf{D}$ and $\mathbf{E}$ are the effective resistance matrices of the ellipsoid head of the swimmer and they are subject to proximity to the solid boundary [46]. Thus, without any other effect in play, one can find the 6-dof rigid-body velocity of the microswimmer, i.e., $[\mathbf{U}\ \mathbf{\Omega}]'$, using (2) and (3) alone for any bacterium cell. And, $\mathbf{S}_{body}$ is the skew-symmetric matrix for the center of volume of the body of the microswimmer. The magnetic effect, $\mathbf{F}_m$ and $\mathbf{T}_m$ in (1), acting on the microswimmer (Fig. 2) is given as follows:

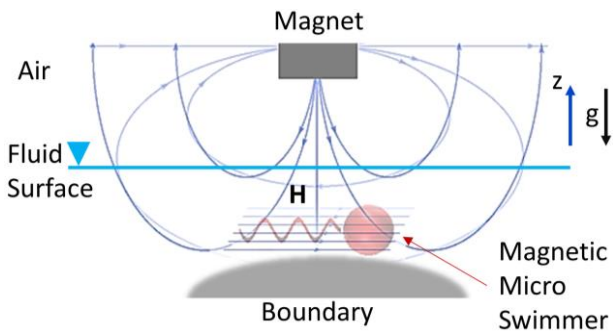

Fig. 2. Micro swimmer fully immersed in the liquid medium while the magnet is right above hanging over at the end-effector of the PPR arm. The electromagnetic field, emanating from the permanent magnet, around the microswimer is depicted with $\mathbf{H} = \mathbf{B}/\mu_0$.

$$\begin{bmatrix} \mathbf{F}_m \\ \mathbf{T}_m \end{bmatrix} = \begin{bmatrix} (\mathbf{m}\cdot\nabla)(\mathbf{R}_{mag}\mathbf{B}) \\ \mathbf{m}\times(\mathbf{R}_{mag}\mathbf{B}) \end{bmatrix}. \quad (4)$$

Here, $\mathbf{m}$ signifies the magnetic moment of the bacterium cell body while $\mathbf{B}$ is the magnetic field generated by the permanent magnet, i.e., $\mathbf{B} = [B_x\ B_y\ B_z]'$, components of which are subject to the position of the microswimmer in the vicinity of the magnet [47] that is being rotated by the third link of the PPR arm and that rotation is taken care of with the help of $\mathbf{R}_{mag}$, i.e., rotation matrix between the swimmer frame and magnet frame. $\mathbf{B}$ in (4) is given as:

$$F_1 = \arctan\frac{(x+a)(y+b)}{(z+c)\left((x+a)^2+(y+b)^2+(z+c)^2\right)^{0.5}}$$

$$F_2 = \arctan\frac{\left((x+a)^2+(y-b)^2+(z+c)^2\right)^{0.5}+b-y}{(z+c)\left((x+a)^2+(y+b)^2+(z+c)^2\right)^{0.5}-b-y}$$

$$B_x = \frac{\mu_0 M}{4\pi}\ln\frac{F_2(-x,y,-z)F_2(x,y,z)}{F_2(x,y,-z)F_2(-x,y,z)}$$

$$B_y = \frac{\mu_0 M}{4\pi}\ln\frac{F_2(-y,x,-z)F_2(y,x,z)}{F_2(y,x,-z)F_2(-y,x,z)} \quad , (5)$$

$$B_z = -\frac{\mu_0 M}{4\pi}\Big[F_1(-x,y,z)+F_1(-x,y,-z) \\ +F_1(-x,-y,z)+F_1(-x,-y,-z) \\ +F_1(x,y,z)+F_1(x,y,-z) \\ +F_1(x,-y,z)+F_1(x,-y,-z)\Big]$$

with $a$, $b$, and $c$ being half-length of the sides to the rectangular prism, measured from the center of volume. Also, $M$ is the total magnetization value of the N52-grade Neodymium magnet [42] and $\mu_0$ is the permeability of the fluidic medium in which the microswimmer is immersed. Next is the weight and buoyancy forces, $\mathbf{F}_g$ and $\mathbf{T}_g$, given as:

$$\begin{bmatrix} \mathbf{F}_g \\ \mathbf{T}_g \end{bmatrix} = \begin{bmatrix} V(\rho_{sw}-\rho_{liquid})\mathbf{R}'_{lab}\mathbf{g} \\ \mathbf{S}_{sw}\mathbf{F}_g \end{bmatrix}, \quad (6)$$

where $V$ is the volume of the head of the microswimmer, $\rho_{sw}$ and $\rho_{liquid}$ are the density of the microswimmer and the liquid medium, respectively, $\mathbf{g}$ is the gravitational attraction in vector form, and $\mathbf{R}_{lab}$ is the rotation matrix from the microswimmer to the lab frame, which in turn rendering gravitational force contributing to 6-dof rigid-body motion as the swimmer can move in all directions freely. Also, $\mathbf{S}_{sw}$ is the skew-symmetric matrix representing the cross product to calculate the torque effect of weight and buoyancy combined. The final component to (1) is the contact force which is not continuous by nature and comes into the picture sporadically:

$$\begin{bmatrix} \mathbf{F}_c \\ \mathbf{T}_c \end{bmatrix} = \begin{bmatrix} \left(k\begin{cases}\delta \Leftarrow \delta <= 0 \\ 0 \Leftarrow \delta > 0\end{cases} + b\begin{cases}d\delta/dt \Leftarrow d\delta/dt > 0 \\ 0 \Leftarrow d\delta/dt <= 0\end{cases}\right)\mathbf{n}_c \\ \mathbf{S}_c\mathbf{F}_c \end{bmatrix}, (7)$$

based on the penalty method modeling [48], where δ stands for the fictitious penetration depth used in simulation with $k$ and $b$ being the spring and damper constants simulating the stiffness and energy absorption at the point of contact. The surface normal at the point of contact is $\mathbf{n}_c = [0\ 0\ 1]'$ (see Fig. 1.) Furthermore, the matrix $\mathbf{S}_c$ is calculated based on the distance to the point of contact from the center of mass of the microswimmer which requires rudimentary meshing along the heady and tail of the microswimmer. This concludes the equation of motion given by (1). Next is the equation of motion for the PPR arm. The Denavit-Hartenberg table [48] used in this study is as follows:

TABLE I. DENAVIT-HARTENBERG TABLE* FOR THE PPR ARM

| Link # | α [rad] | a [m] | d [m] | θ [rad] |
|---|---|---|---|---|
| 1st | −π/2 | 0 | $d_1(t)$ | 0 |
| 2nd | −π/2 | 0 | $d_2(t)$ | π/2 |
| 3rd | 0 | 0 | 0.02 | $\theta_3(t)$ |

*For the orientation given in Fig. 1

The following coupled electromechanical equations representing the dynamics to the PPR arm are [48]:

$$\mathbf{D}(\mathbf{q})\ddot{\mathbf{q}} + \mathbf{C}(\mathbf{q},\dot{\mathbf{q}})\dot{\mathbf{q}} + \mathbf{g}(\mathbf{q}) = \mathbf{K}_m \mathbf{I}_m - \mathbf{r}_m^{-1}\mathbf{T}_{constraint}, \quad (8)$$

$$\mathbf{L}_m \dot{\mathbf{I}}_m + \mathbf{R}_m \mathbf{I}_m = \mathbf{V}_m - \mathbf{K}_b \dot{\mathbf{q}}. \quad (9)$$

Here, the $\mathbf{D}$ is the mass matrix and it is diagonal owing to the dynamics of the PPR arm [48], i.e., $D_{11} = m_1 + m_2 + m_3$, $D_{22} = m_2 + m_3$, $D_{33} = I_{3z}$. Here, $m_{\{1,2,3\}}$ denotes the mass of the links whereas $I_{3z}$ is the area moment of inertia of the third. Likewise, the lack of link length, i.e., the length along the x-axis of each link, denoted by a, eliminates Coriolis and centrifugal forces. Thus, the $\mathbf{C}$ matrix drops out of (8). Finally, none of the joints are doing work against gravity, as can be inspected in Fig. 1, eliminating the vector $\mathbf{g}$ from (8). The vector $\mathbf{q}$ in (8) signifies the generalized coordinates as such $\mathbf{q} = [d_1(t)\ d_2(t)\ \theta_3(t)]' = [y_{sw\text{-}lab}(t)\ x_{sw\text{-}lab}(t)\ \theta_{z\text{-}sw}(t)]'$ in the lab frame owing to effortless inverse kinematics of the PPR arm [48]. Here, the lab frame positions of the microswimmer are given by $[x_{sw\text{-}lab}\ y_{sw\text{-}lab}\ z_{sw\text{-}lab}]' = \mathbf{R}_{lab}[x_{sw}\ y_{sw}\ z_{sw}]'$. Furthermore, $d_1(t) = \kappa_1\theta_1(t)$ and $d_2(t) = \kappa_2\theta_2(t)$ for the associated translation of angular motion to linear motion at the respective joints. The diagonal matrices $\mathbf{K}_m$ and $\mathbf{r}_m$ in (8) hold the torque constants and the gear ratios to the DC-motors, articulating each joint. The vector $\mathbf{T}_{constraint}$ stands for the physical limits exerted on each link should they extend to their designated extrema. The constraint vector elements, with the essential conditions of contact akin to (7), are given as follows:

$$T_{x-\text{constraint}} = -\operatorname{sgn}(d_1)b_d\dot{d}_1(\operatorname{sgn}(d_1\dot{d}_1) > 0) \\ -\operatorname{sgn}(d_1)k_d(d_1 - y_{sw})(d_1 > y_{sw}), \quad (10)$$

$$T_{y-\text{constraint}} = -\operatorname{sgn}(d_2)b_d\dot{d}_2(\operatorname{sgn}(d_2\dot{d}_2) > 0) \\ -\operatorname{sgn}(d_2)k_d(d_2 - x_{sw})(d_2 > x_{sw}), \quad (11)$$

$$T_{z-\text{constraint}} = -\operatorname{sgn}(\theta_3)b_\theta\dot{\theta}_3 \begin{cases} (\operatorname{sgn}(\theta_3\dot{\theta}_3) > 0)^\dagger \\ (\operatorname{sgn}(\theta_3\dot{\theta}_3) < 0)^{\dagger\dagger} \end{cases} \\ -\operatorname{sgn}(\theta_3)k_\theta(\theta_3 - \theta_{z-sw})\begin{cases} (\theta_3 > \theta_{z-sw})^\dagger \\ (-\theta_3 < -\theta_{z-sw})^\dagger \\ (\theta_3 < \theta_{z-sw} - 2\pi)^{\dagger\dagger} \\ (-\theta_3 > \theta_{z-sw} - 2\pi)^{\dagger\dagger} \end{cases}, (12)$$

where the coefficients $b_d$ and $b_\theta$ signify the damping coefficient associated with infinitesimal deformation associated with contact during linear and angular motion, respectively. Akin to the damping coefficient, the spring behavior of the material is signified by $k_d$ and $k_\theta$. The aforementioned properties require tedious computation or careful measurement at the joints of the actual system, however, given that this is a numerical contract to simulate the behavior one can assume that $k_d \approx E\kappa$ with E being Young's modulus and $\kappa$ [m] being the conversion coefficient from meters to radians, i.e., $d_{\{1,2\}} = \kappa\theta_{\{1,2\}}$, and $k_\theta \approx E/(2+2\nu)$ for the third link with ν being the Poisson ratio of the material, which is assumed to be aluminum for all the gears and links except the permanent magnet itself. The respective damping coefficients are designated using the fact that at the instant of contact the link should not undergo an oscillatory motion thus the contact itself should be overdamped as such it is desired to have $\omega_n = (k_\theta/I_{3z})^{0.5}$ and $b_\theta = 2\xi\omega_n I_{3z}$ for the third joint, and $\omega_n = (k_b/m_{\{1,2\}})^{0.5}$ and $b_d = 2\xi\omega_n m_{\{1,2\}}$ for the former two. Hence, the damping ratio, ξ, should be larger than unity. Also, the superscripts $^\dagger$ and $^{\dagger\dagger}$ are used to identify the possible combinations of damper and spring effects in (12).

Finally, the matrix $\mathbf{I}_m$ in (8) is the diagonal matrix of motor currents and it is determined by the solution of (9): $\mathbf{L}_m$, $\mathbf{R}_m$, and $\mathbf{V}_m$ are the diagonal inductance, resistance, and applied voltage matrices to the DC-motors at the joints, respectively, whereas $\mathbf{K}_b$ is the diagonal back-emf constant matrix multiplied by the joint velocities leading to electromechanical coupling. Furthermore, the voltages applied are the output of the 'PID controller – PWM – amplification' sequence. The control law, in effect, sets the duty cycle of the PWM signal subject to the intended direction of motion followed by amplification to the level of the nominal voltage rating of the DC-motors. The control law for the first two joints are:

$$\tau_{\{x,y\}} = K_{p-\{x,y\}}e_{\{x,y\}}(t) \\ + \int K_{i-\{x,y\}}e_{\{x,y\}}(t)\mathrm{d}t + K_{d-\{x,y\}}\dot{e}_{\{x,y\}}(t), \quad (13)$$

$$K_{i-\{x,y\}} = 1/((K_{ia-\{x,y\}}e_{\{x,y\}}(t))^{0.5}+1)$$

with $K_{p-\{x,y\}}$, and $K_{d-\{x,y\}}$ being the proportional and derivative gains for the PID controller whereas the integral gain, $K_{i-\{x,y\}}$, is an adaptive gain based on the initial tuning, i.e., $K_{ia-\{x,y\}}$, and the error itself. The error for the control at the first two joints are formulated as $e_y(t) = d_1(t) - x_{sw}(t)$ and $e_x(t) = d_2(t) - y_{sw}(t)$ owing to the frames depicted in Fig. 1. Next is the third control law, $\tau_{z33}$, employed at the third joint only:

$$\tau_{z31} = K_{p31} e_{\theta-z}(t)$$
$$\tau_{z32} = K_{p32} e_{\dot{\theta}-z}(t) + \int K_{i32} e_{\dot{\theta}-z}(t) \mathrm{d}t, \quad (14)$$
$$\tau_{z33} = K_{p33} e_{I-z}(t) + \int K_{i33} e_{I-z}(t) \mathrm{d}t$$

$$\begin{aligned} e_{\theta-z}(t) &= \theta^{ref}(t) - \theta_{z-sw}(t) \\ e_{\dot{\theta}-z}(t) &= \tau_{z31}(t) - \dot{\theta}_{z-arm}(t), \\ e_{I-z}(t) &= \tau_{z32}(t) - I_{z-m}(t) \end{aligned} \quad (15)$$

$$\begin{aligned} K_{i32} &= 1/((K_{ia32} e_{\dot{\theta}-z}(t))^{0.5}+1) \\ K_{i33} &= 1/((K_{ia33} e_{I-z}(t))^{0.5}+1) \end{aligned}. \quad (16)$$

Here, $I_{z-m}$ is the current of the DC-motor at the third joint. In effect, there are eleven coefficients to determine for the PID control scheme presented here four of which are adaptive [49], [50], i.e., $K_{i-\{x,y\}}$, $K_{i32}$, and $K_{i33}$ in (13) and (14) are changing with the error at any given instance using the initial tunings $K_{ia-\{x,y\}}$, $K_{ia32}$, and $K_{ia33}$ as given by (13) and (16). Here, the orientation of the microswimmer, $\theta_{z-sw}$, rotational velocity of the end effector, $\dot{\theta}_{z-arm}$, and the current flowing through the motor windings, $I_{z-m}$, are all incorporated in the control law while no derivative gain is employed. The PWM signal for the corresponding control signals are generated as follows:

$$\begin{aligned} \chi &= |\tau| / \tau_{max} \\ \psi &= t / f_{PWM} - \mathrm{floor}(t / f_{PWM}), \\ \mathrm{PWM} &= \mathrm{sgn}(\tau) \begin{cases} 1 \Leftarrow \psi < \chi \\ 0 \Leftarrow \psi \geq \chi \end{cases} \end{aligned} \quad (17)$$

with $f_{PWM}$ being the frequency of the PWM signal generation and $\tau_{max}$ signifying the maximum admissible value of control law output over which the value will be automatically saturated, which is set as 100. The PWM signal is then amplified to ±48 V at the end. This is the result of (10)-(17) and constitutes the $\mathbf{V}_m$ in (9).

In effect, equation of motion of the microswimmer, the $T_{z-constraint}$ in the coupled equations of the PPR arm, and the PID control law constitute a layered two-way coupling: the end effector will follow the position of the microswimmer as a set-point to track while the magnetic field of the permanent magnet will exert the magnetic torque on microswimmer altering its direction of swimming while in return the z-orientation of the end-effector is limited by the z-orientation of the microswimmer in the lab frame via the $T_{z-constraint}$. On top of these, a superuser will provide the reference for the heading, $\theta^{ref}(t)$, to complete the loop for which the results are presented in the next section.

## III. RESULTS

The following results are obtained for (i) *M. Gryphiswaldense* species [41] as the microswimmer; (ii) a $0.02 \times 0.02 \times 0.02$ m$^3$ N52- grade Neodymium [42]; (iii) a PPR arm with first two links being rectangular prisms of solid aluminum with a square base of 5 cm × 5 cm and 30 cm in length and the third link being the Neodymium magnet, all articulated by EC 45 Flat (Maxon Group) [51]. The magnet is being moved only along the xy-plane with a fixed z-position above the water surface whereas the microswimmer is fully immersed in the fluidic medium, i.e., water at room temperature, as depicted in Fig. 2.

Here, two distinct scenarios are presented with two different sets of control parameters. They are: (i) $K_{p31} = 15.9256$, $K_{p32} = 14.8509$, $K_{i32} = 3.007$, $K_{p33} = 15.2562$, $K_{i33} = 3.0973$, $K_{p-x} = 12134.9320$, $K_{ia-x} = 759.9446$, $K_{p-y} = 11378.0145$, $K_{ia-y} = 758.8964$, $K_{d-x} = 15.493445$, $K_{d-y} = 14.8192$; and (ii) $K_{p31} = 0.9964$, $K_{p32} = 1.0113$, $K_{i32} = 0.9945$, $K_{p33} = 1.0226$, $K_{i33} = 1.0087$, $K_{p-x} = 809.0164$, $K_{ia-x} = 50.5668$, $K_{p-y} = 758.8198$, $K_{ia-y} = 50.5662$, $K_{d-x} = 1.0113$, $K_{d-y} = 1.0113$. The $f_{PWM}$ is set as 5 kHz and the overdamped condition for constraint torques is imposed by $\xi = 2$. Furthermore, the gear ratio is assumed to be 1:66 at the joints whereas the linear to angular conversion coefficient, $\kappa$, is set as 2 cm. Finally, the magnet is assumed to be hanging 2 cm above the micro swimmer which is initially hovering 0.62 μm, which is the diameter of the cell body in its short axis [41], above the solid boundary. It is noted that, (1), (8), (9), (13) and (14) are solved by forward-integration with fixed time-steps over time [52] under the MATLAB environment running on a 64-bit CPU. And, the performance tests are conducted with the reference of $\theta^{ref}(t) = \pi/36 \sin(0.5\pi t)$ in both cases. In Fig. 3, the motion of the PPR arm and the microswimmer, along the xy-plane, are presented with simulation time. Here, the first control parameter set is used. The reference for the PPR arm is the position of microswimmer while the microswimmer is propelling itself forward via rotation of the helical tail.

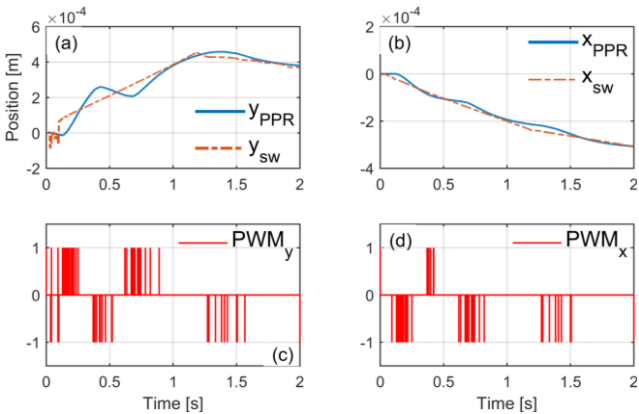

Fig. 3. y- and x-motion of the PPR arm following the micro swimmer, with control coefficient set (i): Respective position control (a), (b); associated PWM signals (c), (d).

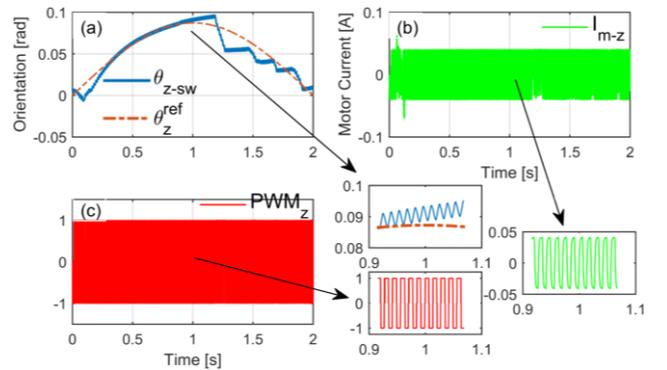

Fig. 4. The micro swimmer heading under PID control, , with control coefficient set (i): Yaw angle (a); instantaneous motor current (b); respective PWM signal (c). Respective insets demonstrating the oscillatory nature are highlighted with arrows. $\theta^{ref}(t) = \pi/36 \sin(0.5\pi t)$

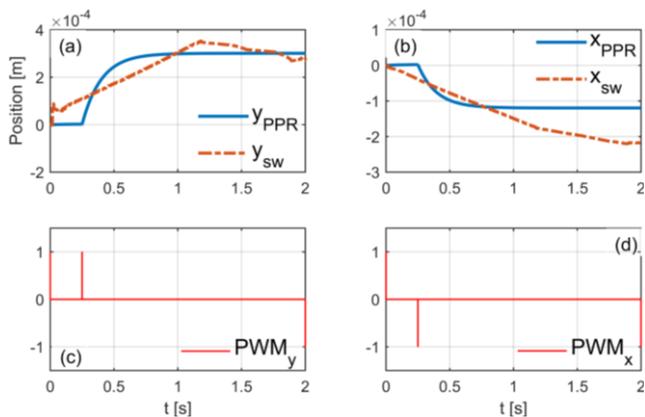

Fig. 5. y- and x-motion of the PPR arm following the micro swimmer, with control coefficient set (ii): Respective position control (a), (b); associated PWM signals (c), (d).

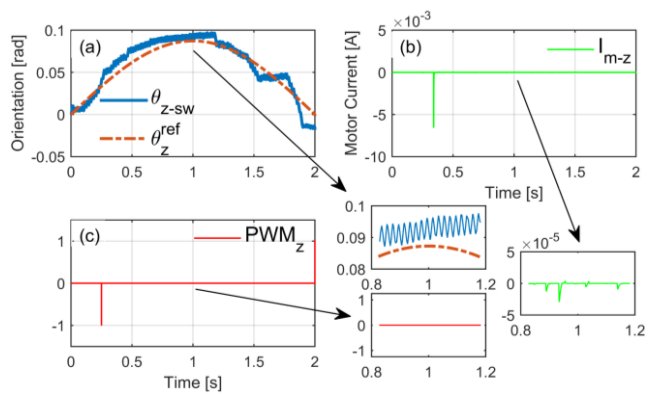

Fig. 6. The micro swimmer heading under PID control, , with control coefficient set (ii): Yaw angle (a); instantaneous motor current (b); respective PWM signal (c). Respective insets demonstrating the oscillatory nature are highlighted with arrows. $\theta_z^{ref}(t) = \pi/36\sin(0.5\pi t)$

The maximum set point tracking error is on the order of $O(-5)$ and the performance of the control law improves over time (Fig. 3(a) and Fig. 3(b)) which also reflects on the PWM signal generation towards the end of the period (Fig. 3(c) and Fig. 3(d)). Notice that the PWM signal changes direction as the control law predicts reversal in direction of motion at the respective joints of the PPR arm. In Fig. 4, the yaw angle, $\theta_{z\text{-}sw}$, is given. The microswimmer is following the reference externally provided under the influence of the magnetic torque. The reference imposed is The maximum set point tracking error is on the order of $O(-2)$ and the control law struggles to keep the heading of the swimmer accurate especially during the second half of the simulation (Fig. 4(a)). The motor current for the third axis, $I_{m-z}$, oscillates between ±0.4 A constantly (Fig. 4(b)) and the associated PWM signal also shows the same behavior (Fig. 4(c)). Insets are provided to emphasize the oscillating dynamics of the microswimmer and the control law.

In Fig. 5, the setpoint tracking performance of the PPR arm is demonstrated with the second set of control parameters. The maximum set point tracking error, with a notable increase as opposed to the previous example, is on the order of $O(-4)$ on both axes (Fig. 5(a) and Fig. 5(b)). The PWM signals indicate that the control law did not interfere with the position at the first two joints of the PPR arm frequently (Fig. 5(c) and Fig5. (d)). In Fig. 6, depicts the microswimmer following the same yaw angle reference: the performance of the control law improves and deteriorates sporadically (Fig. 6(a)). The motor current and PWM signals associated with this performance demonstrate a very effortless control (Fig. 6(b) and Fig. 6(c)) as such it renders the previous performance (Fig 4) excessive on its part. Thus, it can arguably be deduced that neither example is optimum; however, the optimality of the set of control parameters depend on the accuracy and precision requirement of the desired task. Finally, insets represent that the oscillatory behavior of microswimmer is expected regardless of the control effort, i.e., while the PWM signal and the motor current are being zero.

## IV. CONCLUSIONS

The detailed mathematical model takes care of a variety of physical stimuli to capture the real condition as much as possible. It is demonstrated that the heading of the microswimmer can be controlled with relatively high accuracy without the same level of precision on tracking the position of the microswimmer. Nonetheless, bilateral control of two robots is achieved with acceptable set point tracking errors with two parameter sets. The *M. Gryphiswaldense* species in this study might not reflect the ability or limitations of all the magnetically controlled bacteria, of natural species or originate in the laboratory via scientific intervention; however, it sets an example on how these cells could be controlled. More detailed modeling on the tactic behavior, e.g., biological responses, would bring the performance of adaptive control further in focus. The method could arguably be useful in deep tissue provided the field is strong enough and the overall system is fast enough although degradation is expected nonetheless without additional magnets.